\documentclass[10pt,twocolumn,letterpaper]{article}

\usepackage{iccv}
\usepackage{times}
\usepackage{epsfig}
\usepackage{graphicx}
\usepackage{amsmath}
\usepackage{amssymb}
\usepackage{algorithmic}
\usepackage{algorithm}
\usepackage{multirow}
\usepackage{color}

\usepackage[subrefformat=parens]{subcaption}

\usepackage[breaklinks=true,bookmarks=false]{hyperref}

\iccvfinalcopy 

\def\iccvPaperID{****} 

\ificcvfinal\fi

\begin{document}

\title{MSE Loss with Outlying Label for Imbalanced Classification}

\author{Sota Kato\\
Meijo University\\
{\tt\small 150442030@ccalumni.meijo-u.ac.jp}
\and
Kazuhiro Hotta\\
Meijo University\\
{\tt\small kazuhotta@meijo-u.ac.jp}
}

\maketitle
\ificcvfinal\thispagestyle{empty}\fi

\begin{abstract}
In this paper, we propose mean squared error (MSE) loss with outlying label for class imbalanced classification. 
cross entropy (CE) loss, which is widely used for image recognition, is learned so that the probability value of true class is closer to one by back propagation. 
However, for imbalanced datasets, the learning is insufficient for the classes with a small number of samples. 
Therefore, we propose a novel classification method using the MSE loss that can be learned the relationships of all classes no matter which image is input. 
Unlike CE loss, MSE loss is possible to equalize the number of back propagation for all classes and to learn the feature space considering the relationships between classes as metric learning. 
Furthermore, instead of the usual one-hot teacher label, we use a novel teacher label that takes the number of class samples into account. 
This induces the outlying label which depends on the number of samples in each class, and the class with a small number of samples has outlying margin in a feature space. 
It is possible to create the feature space for separating high-difficulty classes and low-difficulty classes. 
By the experiments on imbalanced classification and semantic segmentation, we confirmed that the proposed method was much improved in comparison with standard CE loss and conventional methods, even though only the loss and teacher labels were changed.
\end{abstract}

\section{Introduction}
\begin{figure}[t]
\begin{center}
\includegraphics[scale=0.22]{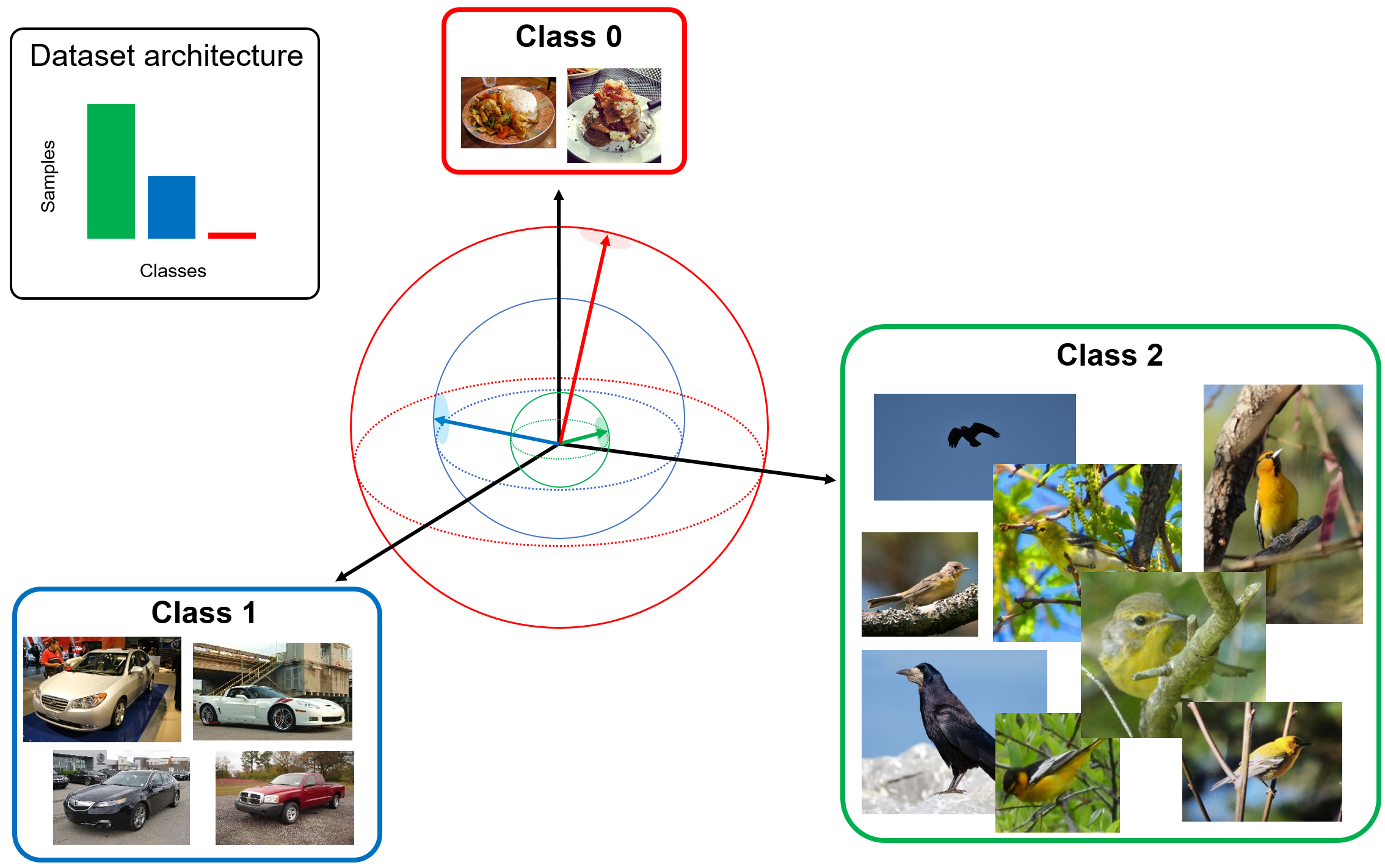}
\end{center}
\caption{Overview of the proposed method. In the imbalanced dataset which consists of three categories, it is possible to create the feature space that the category with a large number of samples is placed inlying and the category with a small number of samples is placed outlying using the MSE loss with outlying label.}
\label{fig:long}
\label{fig:onecol}
\end{figure}
In recent years, Convolutional Neural Networks (CNNs) have been known to produce high accuracy for image recognition and are widely used for training on large datasets\cite{kuznetsova2020open, lin2014microsoft, yungbig}. 
However, there are often differences in the number of samples in each class when CNNs are applied to real-world datasets\cite{liu2019large,van2018inaturalist}. 
In image classification, there is the problem that images belonging to a class are accounted for the majority of datasets and images belonging to remaining classes are few\cite{bergmann2019mvtec, real2017youtube}. 
When we train a classifier on the dataset with an imbalanced sample ratio between classes, the accuracy of the class with a small number of samples will be extremely low. 
Improving the classification accuracy on imbalanced datasets is an important problem to apply deep learning to real-world datasets. 

To solve this problem, re-sampling methods are often used as countermeasure methods\cite{chawla2009data, chawla2002smote}. 
However, although these methods are changed the apparent number of data, they have not been changed the essential number of data itself. 
Re-sampling methods do not solve the fundamental problem of class imbalanced datasets.
To improve the accuracy, it is important to perform efficient learning with a limited number of samples, and loss functions has been improved in recent researches\cite{cao2019learning,cui2019class,li2019gradient,lin2017focal,ren2020balanced,sinha2020class,tan2020equalization,yang2020rethinking}. 
However, almost of the conventional researches are based on the cross entropy (CE) loss which is widely used for image recognition, and we consider that there is a better loss than CE loss. 
Therefore, we focus on new loss function that does not rely on CE loss and is effective for class imbalanced datasets. 

From the issues of CE loss, we consider that the cause of low accuracy of some classes in class imbalanced datasets is the bias of the number of back propagation, and we propose a new classification method using the loss function based on MSE loss as a way to achieve better learning.
MSE loss is possible to equalize the number of back propagation for all classes and to learn the feature space considering the relationships between classes.  
Furthermore, we proposed to use outlying labels that take the number of samples in each class into account.
Figure 1 shows the overview of our method using the MSE loss with outlying class labels.
The outlying class label is defined according to the number of samples in each class.
By using outlying labels, the class with a small number of samples is placed the farthest point and the class with the largest number of samples is placed the nearest point.  
The farther prediction point is, the larger loss has. 
Thus, the network is enhanced by giving a larger loss to the difficult classes with few samples.

We conducted two types of experiments on imbalanced classification and semantic segmentation. 
From the experimental results, we confirmed that the proposed method was much improved in comparison with CE loss and conventional methods for imbalanced datasets, even though we changed only the loss function and teacher labels.

This paper is organized as follows. Section 2 describes related works. Section 3 explains the details of the proposed method. We describe the datasets and evaluation method in section 4, Section 5 shows the experimental results. Finally, we describe our summary and future works in section 6.

\section{Related works}

This section describes conventional two kinds of works for class imbalance datasets; re-sampling and cost-sensitive learning.

\subsection{Re-samplig}

There are three kinds of re-sampling methods; wrapping, over sampling and under sampling\cite{summers2019improved,takahashi2019data,wang2017effectiveness,zhang2017mixup,zhong2020random}. 
Wrapping is the method that applied geometric transformations to images included in a minority class.
A single image can be processed by rotation, affine transformation, gaussian noise, changing of luminance value and random erasing\cite{zhong2020random} to produce the images with different characteristics from the original image.

The method for generating new data from existing data is called oversampling\cite{summers2019improved,takahashi2019data,wang2017effectiveness,zhang2017mixup}.
Sample pairing\cite{wang2017effectiveness} is to enhance training data by generating the averaged image of two images randomly selected from the original dataset. 
Even though sample pairing\cite{wang2017effectiveness} combines two images linearly, there is also a non-linear method\cite{summers2019improved}. 
In sample pairing, the image averaged of two images is generated, but in mixup\cite{zhang2017mixup}, the ratio of two images is obtained from the sampling of the beta distribution. 
There is also random image cropping and patching method\cite{takahashi2019data}, which is the method for randomly cropping and combining the regions using multiple images.
Undersampling is to sample from a large number of classes in order to fit the number of samples from the small number of classes\cite{chawla2009data}.
However, this method is not widely used because it reduces whole number of data.

\subsection{Cost-sensitive learning}

Cost-sensitive learning techniques try to be learned by imposing a large cost on unacceptable errors and a small cost on acceptable errors. 
There are two main methods; the first one is based on the cost of the distribution of class labels\cite{badrinarayanan2017segnet,cao2019learning,cui2019class,tan2020equalization,yang2020rethinking} and the other is based on the cost of the probability of each sample\cite{li2019gradient,lin2017focal}.

Class-wise weight\cite{badrinarayanan2017segnet,cui2019class} is to take into account the distribution of class labels. 
This method calculated the weights based on the number of samples in each class and weighted the loss function to increase the penalty for making a mistake in a class with a small number of samples. 
Since imbalanced datasets have a bias in learning among classes due to the difference in the number of samples, it is possible to equalize the balance of learning among classes using class-wise weight. 
Badrinarayana et al. proposed a class weight using median sample for semantic segmentation\cite{badrinarayanan2017segnet}. 
Weights become smaller for the classes with more samples and larger for the classes with fewer samples. 
By multiplying these weights by loss function, it is possible to perform the learning that takes into account the class balance.  
Class-balanced loss using effective number\cite{cui2019class} was proposed.
It calculated the effective number of each class and used it to assign weights to samples. 
As the number of samples increases, the limit of the amount of information that the model can extract from the data decreases because there is the overlap of information among the data. 
Cao et al. proposed a loss function with the margin that takes into account the label distribution\cite{cao2019learning}. 
This margin is obtained from the total number of data divided by the 1/4 power of the number of samples for each class. 
By using this loss function, it is possible to obtain a large margin for the class with a small number of samples and a small margin for the class with a large number of samples

Focal loss\cite{lin2017focal} is the method that is used the cost composed of the probability of each sample instead of fixed weights.
It is dynamically scaled CE loss by reducing the penalty for easy samples and increasing the penalty for difficult samples. 
Focal loss is allowed us to learn adaptive weights for each sample unlike class-wise weight.
There were the methods \cite{sinha2020class, tan2020equalization} that each class with different number of samples is weighted as Focal loss.
Equalization loss\cite{tan2020equalization} is the loss function for long-tailed object recognition.
This loss protects the learning of rare categories from being disadvantageous during the network parameter updating, and the model is able to learn better discriminative features for objects of rare classes.
Sinha et al. proposed a class-wise difficulty balanced loss for solving class imbalanced classification\cite{sinha2020class}. 
They use difficulty levels of each class and the weights which dynamically change are used as the difficulty levels for the model. 

Many conventional methods were proposed, and almost of all conventional methods are balanced learning between easy classes and difficult classes while changing the weights of CE loss. 
However, we consider that the optimal weight is different for each dataset and it is very difficult choosing the best weight.
In actual, no method has been determined to be effective for any imbalanced datasets.
We need to propose new loss and solve this problem fundamentally.

\section{Proposed method}
\begin{figure}[t]
    \centering
    \includegraphics[scale=0.25]{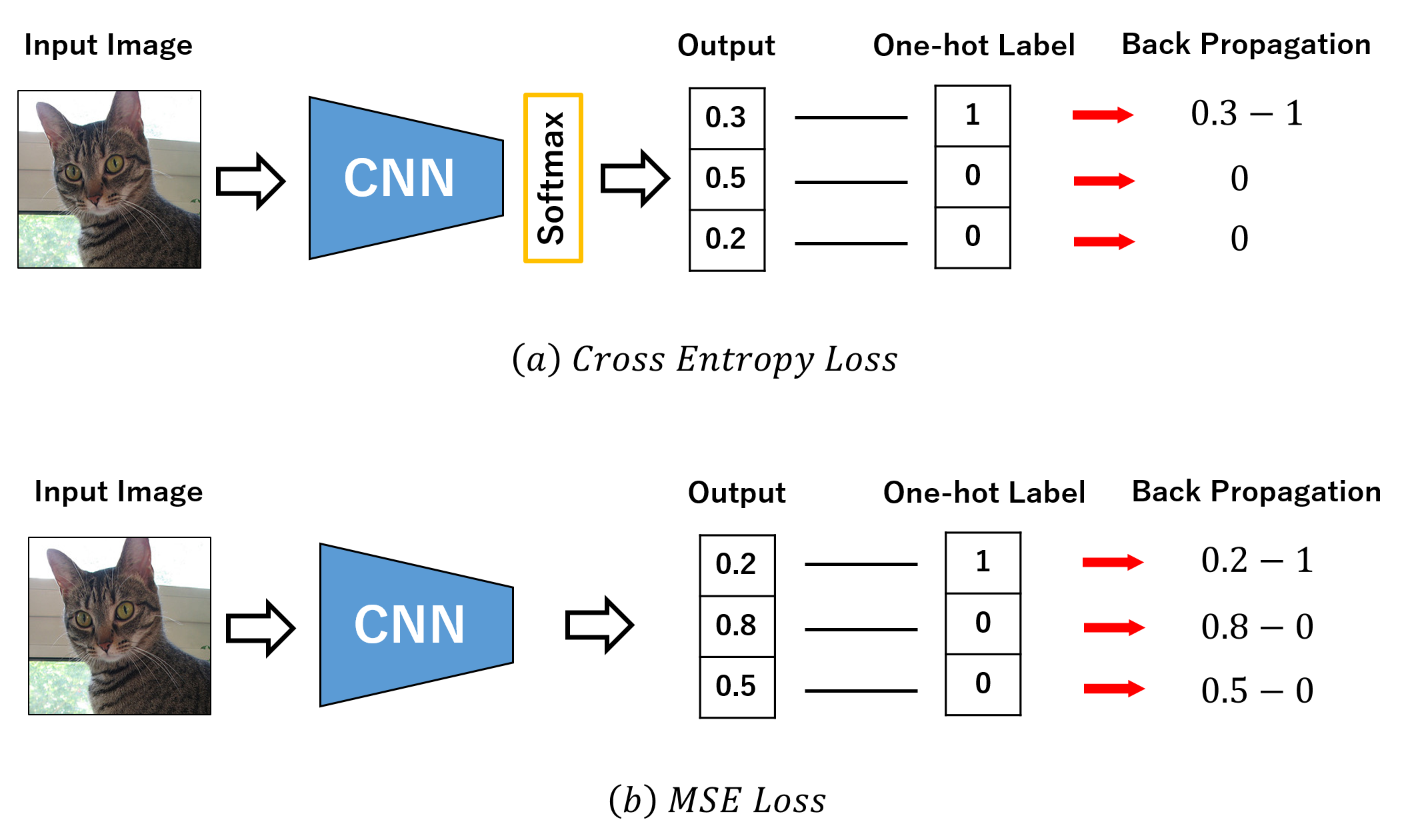}
    \caption{Difference of CE loss and MSE loss. The top row is the overview of training by CE loss and the bottom row is the training by MSE loss.}
    \label{fig:my_label1}
\end{figure}

This section describes our proposed method using the MSE loss instead of CE loss for imbalanced datasets. 
MSE loss is often used in regression problems to predict actual numbers, and it has rarely been used for multi-class classification. 
However, we consider that MSE loss is better than CE loss for learning on class imbalanced datasets. 
In the following sections, we show why MSE loss is superior to CE loss for the datasets. 

Section 3.1 describes the problems of CE loss on imbalanced datasets, and Section 3.2 describes the proposed method based on MSE loss. 

\subsection{Issues of CE loss for imbalanced dataset}
\begin{figure}[t]
    \centering
    \begin{tabular}{cc}
      \begin{minipage}{0.45\hsize}
        \centering
        \includegraphics[scale=0.26]{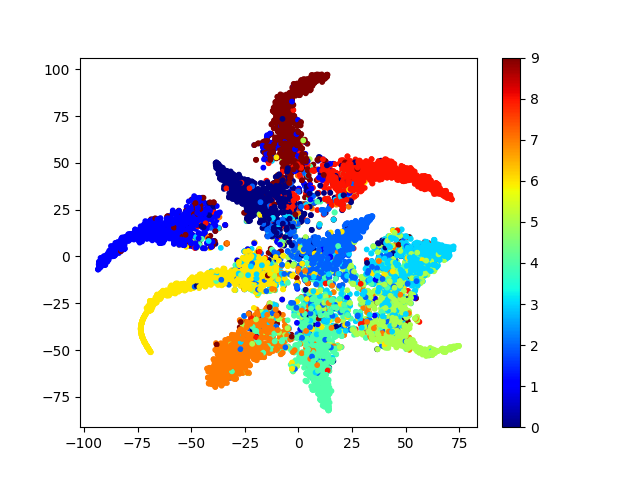}
    \end{minipage}%
    \begin{minipage}{0.45\hsize}
        \centering
        \includegraphics[scale=0.26]{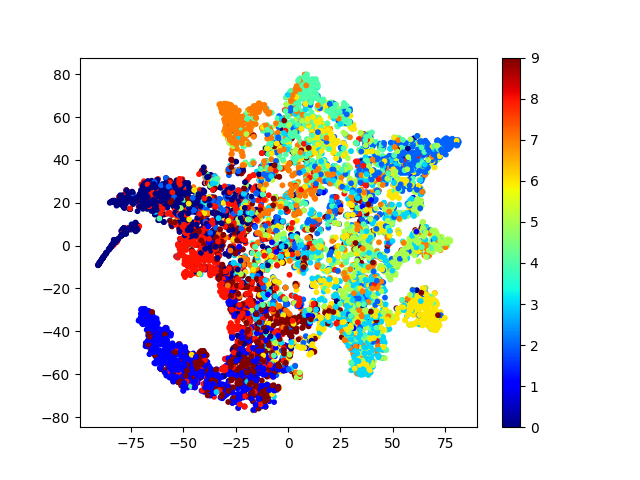}
    \end{minipage}%
    \end{tabular}
    \caption{Visualization of penultimate layer’s activations in the ResNet34 on the CIFAR-10 dataset by t-SNE\cite{van2008visualizing}. 
    The left column shows the visualization of test images in the original CIFAR-10 dataset, and the right column shows the visualization for the long-tailed type of CIFAR-10. In the dataset, the larger the class number, the smaller the number of samples in the class.
    }
\end{figure}

Equation (1) shows the CE loss and equation (2) shows the update equation for network parameters. 
\begin{eqnarray}
  CE Loss & = & \sum_{k=1}^K t_k\log y_k \\
  \frac{\partial CE Loss}{\partial a_i} & = & y_k-t_k
\end{eqnarray}
where $K$ is the number of classes in the dataset, $t_k$ is a teacher label for the class $k$, $a_i$ is the $i$-th element of $\textbf{a}$ which is an output vector of deep neural network, $y_k$ is the probability value after a softmax function as $y_k=\frac{e^{a_i}}{e^{a_i}+ \sum_{j\neq i}^J e^{a_j}}$.
As shown in equation (2), network parameters are learned to minimize the difference between the output probability value and the teacher label.
However, since the teacher label by CE loss is represented as a one-hot vector that the dimension of the true class is 1 and the other dimensions are 0, the loss between the probability value of false class and teacher label is not used in equation (2).
In other words, it can be seen that learning with CE loss is only brought the probability value of true class closer to one as shown in equation (3) when one-hot vector is used as teacher label.
\begin{eqnarray}
  \frac{\partial CE Loss}{\partial a_i} = \left\{ \begin{array}{ll}
    y_k-1 & (k=i) \\
    0  & (k\neq i)
  \end{array} \right.
\end{eqnarray}

Figure 2 shows the difference of CE loss and MSE loss in back propagation.
This is not a problem when there is no class imbalance in a dataset.
However, since the number of error back propagation of each class is different in class imbalanced dataset, the accuracy of the class with a few samples becomes low.

We also check the feature space in the network trained on the imbalanced CIFAR-10 using CE loss. The feature space obtained by t-SNE\cite{van2008visualizing} is shown in Figure 3. In the case of original CIFAR-10 as shown in left column, the feature space is created where each class is independent. However, in the case of imbalanced CIFAR-10 as shown in right column, the feature points of the class with the largest number of samples (blue) and the class with the smallest number of samples (red) overlap. 
From the above results, the bias in the number of back propagation can be solved by learning the predictions of all the classes in the dataset even when the samples belonging to a certain class are input. Feature overlapping of different classes can be solved by learning to separate the distances between features of different classes.

\subsection{Learning method using MSE loss}
\begin{figure}[t]
    \centering
    \includegraphics[scale=0.25]{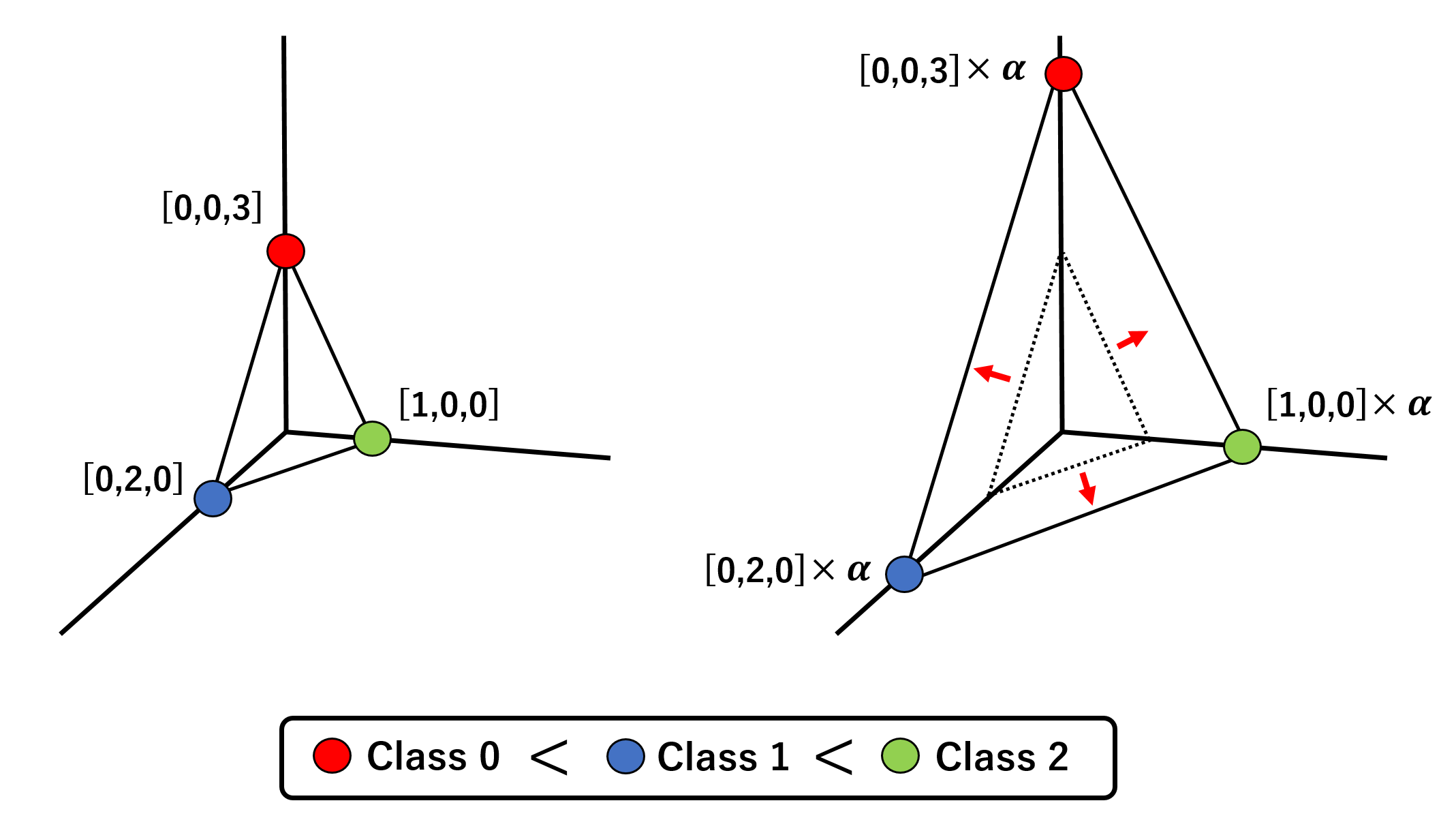}
    \caption{Overview of the proposed method using outlying label. In the case of three class classification, class 0 has the smallest number of samples (red point), class 1 has middle number of samples (blue point) and class2 is the largest number of samples (green point).  Outlying label has different prediction point for each class, and class 0 has the farthest label. By changing $\alpha$, the difference between prediction points for each class becomes large.}
    \label{fig:my_label1}
\end{figure}
Equation (4) shows the MSE loss and equation (5) shows its update equation of the network parameters when one hot vector is used as teacher label.
\begin{eqnarray}
  MSE Loss & = & \sum_{k=1}^K \frac{1}{2}\|a_k-t_k\|^2 \\
  \frac{\partial MSE Loss}{\partial a_i} & = & \left\{ \begin{array}{ll}
    a_i-1 & (k=i) \\
    a_i-0  & (k\neq i)
  \end{array} \right.
\end{eqnarray}
where $K$ is the number of classes in the dataset, $t_k$ is the teacher label, and $a_k$ is the $k$-th output of the network. 
Equation (5) can be expressed by the difference between the output probability and the teacher label in the same way as CE loss in equation (2). 
However, when we compare equation (3) with equation (5), MSE loss is learned to closer the feature vector of the false class to zero. 
Unlike CE loss, MSE loss is possible to equalize the number of back propagation in all classes.
At the time of inference, we only need to determine which vector is the largest and we can classify a test image in the same way as conventional methods using CE loss.
In addition, MSE loss learns the feature space considering the relationships between classes as metric learning because MSE loss is learning to be closer to the true class and be far from the false class.

\subsection{MSE loss with outlying label}
We consider that there would be an appropriate prediction point for each class when we deal with class imbalanced datasets because the features of the class with the largest number of samples will be closer to the features of the other classes.
Therefore, we use outlying label which changes the length of the vectors corresponding to the true class as shown in Figure 4 instead of using standard one-hot teacher labels. 
In the proposed outlying label, we can select the class with the largest number of samples as the standard class. It is possible to learn to place the class with the largest number of samples as a reference and place the class with smaller number of samples farther away. 
The outlying class label is set according to the reciprocal of training samples in each class, and the class with the smallest number of samples is placed the farthest point.
Since it is easy to predict the closer prediction points and difficult to predict the far prediction points, the network is enhanced by giving a larger loss to the difficult classes with few samples. 

Algorithm 1 shows how to set the outlying label, where $k$ is class, $C_k$ is the number of training images in each class, $L_k$ is the one-hot type teacher label, $\alpha$ is a hyper parameter, $S_C$ is the class index vector lined up in order of the large number of samples and $NewLabel$ is the outlying label.
The number of training samples for each class is sorted in descending order and outlying label is created in such a way that the value corresponding to true class in the one-hot teacher label is increased by adding $N+1$ according to the descending order of the number of class samples.
In this way, network is learned that classes with small samples can be placed far from the origin in the feature space, while classes with large samples can be placed closer to the origin. 
Furthermore, by multiplying the outlying label by higher $\alpha$, the difference between the classes with few and many samples can be widened. 
The $\alpha$ is determined by the validation dataset. 

\begin{algorithm}[t]
    \caption{Outlying label}
    \label{alg1}
    \begin{algorithmic}[1]
    \REQUIRE $C=\{C_1,C_2,...C_k\}, L=\{L_1,L_2,...L_k\}, \alpha$ 
    \STATE $S_C = argsort(C), N=0$
    \FOR{$ len(S_C) $}
    \STATE $NewLabel_{S_C} \leftarrow L_{S_{C}}\times(N+1)$
    \STATE $N \leftarrow N+1$
    \ENDFOR
    \STATE $NewLabel \leftarrow NewLabel \times \alpha$
    \ENSURE $NewLabel$
    \end{algorithmic}
\end{algorithm}

\begin{figure}[t]
    \centering
    \begin{tabular}{cc}
      \begin{minipage}{0.45\hsize}
        \centering
        \includegraphics[scale=0.2]{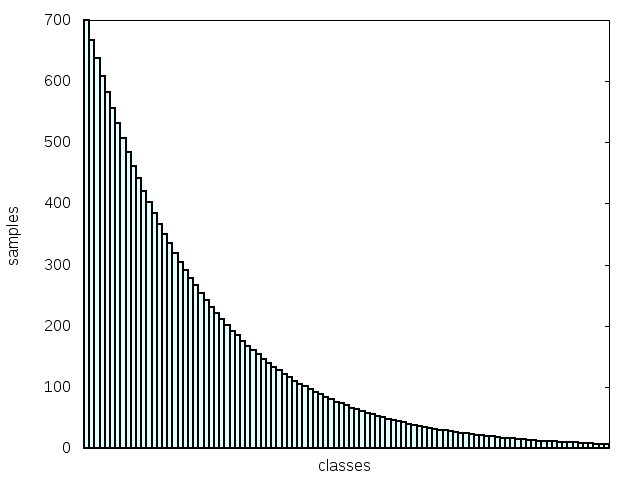}
    \end{minipage}%
    \begin{minipage}{0.45\hsize}
        \centering
        \includegraphics[scale=0.2]{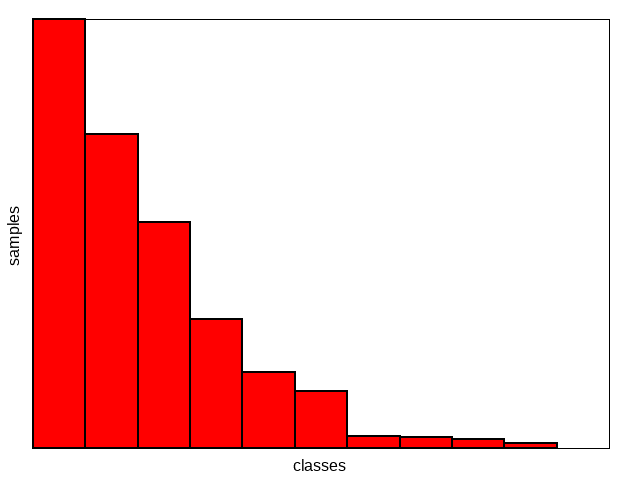}
    \end{minipage}%
    \end{tabular}
    \caption{The distribution of training samples in two datasets. The left column shows the distribution of long-tailed Food-101 dataset and right column shows the distribution of the CamVid dataset}
\end{figure}

\begin{table*}[t]
    \centering
    \caption{Top-1 accuracy on imbalanced CIFAR-10.}
    \scalebox{0.75}{
    \begin{tabular}{c||cccc||cccc} \hline
    Imbalance Type & \multicolumn{4}{c||}{Long-tailed Type} & \multicolumn{4}{c}{Step Type} \\ 
    \hline
    Imbalance Ratio & 5 & 10 & 50 & 100 & 5 & 10 & 50 & 100 \\ 
    \hline
        CE Loss(Vanilla)&68.78(±0.12)&64.25(±0.30)&50.29(±0.89)& 44.50(±0.53)&67.48(±0.15)&61.36(±0.52)&47.62(±1.10)&43.87(±0.68) \\
        Focal Loss\cite{lin2017focal}&67.80(±0.82)&63.83(±0.50)&48.53(±0.43)&43.78(±1.05)&67.58(±0.26)&60.66(±0.33)&46.58(±1.08)&43.49(±0.55) \\
        LDAM Loss\cite{cao2019learning}&68.69(±0.62)&64.32(±0.79)&49.97(±0.32)&43.63(±1.28)&68.24(±0.23)&61.77(±1.32)&47.52(±0.75)&43.09(±1.34) \\
        CDB-CE Loss\cite{sinha2020class}&67.23(±0.46)&60.56(±1.42)&\textbf{52.21(±1.61)}&43.97(±0.81)&65.32(±1.66)&58.45(±1.75)&\textbf{48.23(±1.60)}&43.65(±0.65) \\
        Equalization Loss\cite{tan2020equalization}&69.11(±0.12)&63.84(±0.81)&49.85(±0.07)&44.58(±0.72)&67.89(±0.49)&61.78(±1.50)&48.03(±1.03)&43.20(±0.20) \\
    \hline
        MSE Loss(Vanilla)&66.95(±1.20)&62.03(±0.39)&47.76(±0.41)&42.86(±1.13)&66.63(±0.20)&60.06(±0.17)&47.26(±0.24)&41.90(±0.64) \\
        MSE Loss with OL($\alpha=5$)&\textbf{69.26(±0.47)}&\textbf{64.60(±0.58)}&50.91(±0.29)& \textbf{46.99(±1.07)}&\textbf{68.33(±0.51)}&\textbf{62.17(±0.71)}&47.69(±1.43)&\textbf{44.53(±1.39)} \\
    \hline
    \end{tabular}
    }
\end{table*}
\begin{table*}[t]
    \centering
    \caption{F-measure on imbalanced CIFAR-10}
    \scalebox{0.75}{
    \begin{tabular}{c||cccc||cccc} \hline
    Imbalance Type & \multicolumn{4}{c||}{Long-tailed Type} & \multicolumn{4}{c}{Step Type} \\ 
    \hline
    Imbalance Ratio & 5 & 10 & 50 & 100 & 5 & 10 & 50 & 100 \\ 
    \hline
        CE Loss(Vanilla)&68.94(±0.11)&64.14(±0.34)&47.23(±1.02)&39.25(±0.59)&67.15(±0.24)&60.18(±0.72)&41.63(±1.84)&35.72(±0.87) \\
        Focal Loss\cite{lin2017focal}&67.96(±0.75)&63.67(±0.49)&45.35(±0.64)&38.79(±1.14)&67.40(±0.22)&59.44(±0.29)&40.72(±1.18)&35.42(±0.67) \\
        LDAM Loss\cite{cao2019learning}&68.86(±0.54)&64.20(±0.78)&47.05(±0.37)&38.36(±0.94)&68.08(±0.28)&60.94(±1.70)&42.13(±1.02)&34.58(±2.42) \\
        CDB-CE Loss\cite{sinha2020class}&66.80(±0.50)&60.36(±1.55)&\textbf{49.75(±2.48)}&39.33(±1.09)&64.77(±2.35)&56.92(±2.04)&\textbf{42.98(±2.47)}&34.96(±1.16) \\
        Equalization Loss\cite{tan2020equalization}&69.37(±0.06)&63.61(±0.90)&46.72(±0.19)&40.01(±0.88)&67.65(±0.59)&60.87(±1.79)&42.84(±1.72)&34.27(±0.41) \\
    \hline
        MSE Loss(Vanilla)&66.82(±1.18)&61.72(±0.41)&44.20(±0.61)&37.70(±1.37)&66.37(±0.16)&58.80(±0.26)&41.39(±0.59)&32.95(±0.43) \\
        MSE Loss with OL($\alpha=5$)&\textbf{69.40(±0.53)}&\textbf{64.51(±0.67)}&48.13(±0.36)& \textbf{42.89(±2.07)}&\textbf{68.32(±0.54)}&\textbf{61.23(±0.79)}&42.03(±2.57)&\textbf{36.51(±2.58)} \\
    \hline
    \end{tabular}
    }
\end{table*}
\begin{table*}[t]
    \centering
    \caption{Top-1 accuracy on imbalanced CIFAR-100.}
    \scalebox{0.75}{
    \begin{tabular}{c||cccc||cccc} \hline
    Imbalance Type & \multicolumn{4}{c||}{Long-tailed Type} & \multicolumn{4}{c}{Step Type} \\ 
    \hline
    Imbalance Ratio & 5 & 10 & 50 & 100 & 5 & 10 & 50 & 100 \\ 
    \hline
        CE Loss (Vanilla)&36.02(±0.79)&32.76(±0.25)&24.46(±0.34)&21.78(±0.51)&35.68(±0.14)&31.90(±0.68)&26.53(±0.26)&25.91(±0.04) \\
        Focal Loss\cite{lin2017focal}&35.77(±0.54)&31.52(±0.57)&23.97(±0.34)&21.67(±0.41)&34.83(±0.40)&30.96(±0.46)&26.37(±0.35)&26.15(±0.47) \\
        LDAM Loss\cite{cao2019learning}&27.23(±0.43)&23.72(±0.49)&16.98(±2.59)&17.67(±0.19)&28.67(±1.75)&25.68(±1.57)&23.09(±0.59)&21.31(±0.51) \\
        CDB-CE Loss\cite{sinha2020class}&35.97(±0.35)&32.86(±0.71)&24.68(±0.49)&21.86(±0.24)&34.54(±0.64)&31.67(±0.57)&26.76(±0.29)&26.44(±0.17) \\
        Equalization Loss\cite{tan2020equalization}&36.05(±1.04)&31.70(±0.68)&25.04(±0.36)&22.05(±0.25)&35.07(±0.28)&31.19(±0.14)&26.58(±0.13)&25.92(±0.27) \\
    \hline
        MSE Loss(Vanilla)&29.65(±0.90)&25.27(±0.93)&19.89(±0.33)&18.48(±0.29)&30.14(±1.06)&26.20(±0.29)&23.41(±0.87)&24.37(±0.58) \\
        MSE Loss with OL($\alpha=1$)&\textbf{38.58(±0.22)}&\textbf{35.35(±0.07)}&\textbf{28.96(±0.37)}& \textbf{26.65(±0.74)}&\textbf{39.59(±1.56)}&\textbf{33.94(±2.49)}&\textbf{29.25(±0.42)}&\textbf{29.18(±1.03)} \\
    \hline
    \end{tabular}
    }
\end{table*}
\begin{table*}[t]
    \centering
    \caption{F-measure on imbalanced CIFAR-100.}
    \scalebox{0.75}{
    \begin{tabular}{c||cccc||cccc} \hline
    Imbalance Type & \multicolumn{4}{c||}{Long-tailed Type} & \multicolumn{4}{c}{Step Type} \\ 
    \hline
    Imbalance Ratio & 5 & 10 & 50 & 100 & 5 & 10 & 50 & 100 \\ 
    \hline
        CE Loss (Vanilla)&35.11(±0.80)&31.30(±0.32)&21.43(±0.41)&17.92(±0.54)&33.71(±0.21)&28.27(±0.71)&18.99(±0.27)&18.12(±0.03) \\
        Focal Loss\cite{lin2017focal}&34.95(±0.57)&30.16(±0.48)&20.91(±0.48)&17.92(±0.35)&32.83(±0.41)&27.57(±0.57)&18.91(±0.20)&18.10(±0.37) \\
        LDAM Loss\cite{cao2019learning}&26.27(±0.38)&22.46(±0.45)&14.53(±2.25)&14.37(±0.08)&26.83(±1.74)&22.52(±1.51)&16.61(±0.51)&14.83(±0.36) \\
        CDB-CE Loss\cite{sinha2020class}&34.92(±0.34)&31.41(±0.57)&21.60(±0.30)&18.11(±0.37)&32.42(±0.62)&27.85(±0.62)&19.29(±0.19)&18.34(±0.20) \\
        Equalization Loss\cite{tan2020equalization}&35.22(±0.98)&30.23(±0.67)&22.08(±0.22)&18.41(±0.27)&33.23(±0.17)&27.65(±0.07)&19.20(±0.26)&17.94(±0.27) \\
    \hline
        MSE Loss(Vanilla)&28.86(±0.88)&23.69(±0.84)&17.03(±0.38)&14.47(±0.85)&28.03(±1.11)&22.80(±0.10)&16.66(±0.72)&16.68(±0.45) \\
        MSE Loss with OL($\alpha=1$)&\textbf{38.49(±0.79)}&\textbf{34.64(±0.79)}&\textbf{26.59(±0.79)}& \textbf{23.24(±0.79)}&\textbf{37.05(±1.76)}&\textbf{31.09(±2.55)}&\textbf{20.50(±0.75)}&\textbf{20.19(±0.63)} \\
    \hline
    \end{tabular}
    }
\end{table*}

\begin{figure*}[t]
    \centering
    \begin{tabular}{cccc}
      \begin{minipage}{0.24\hsize}
        \centering
        \includegraphics[scale=0.3]{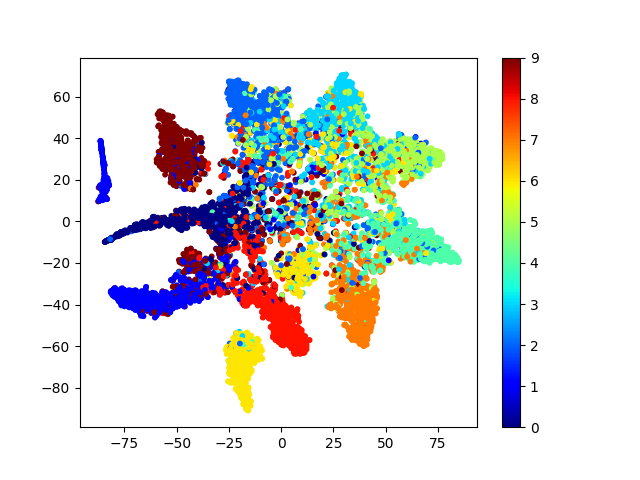}
    \end{minipage}%
    \begin{minipage}{0.24\hsize}
        \includegraphics[scale=0.3]{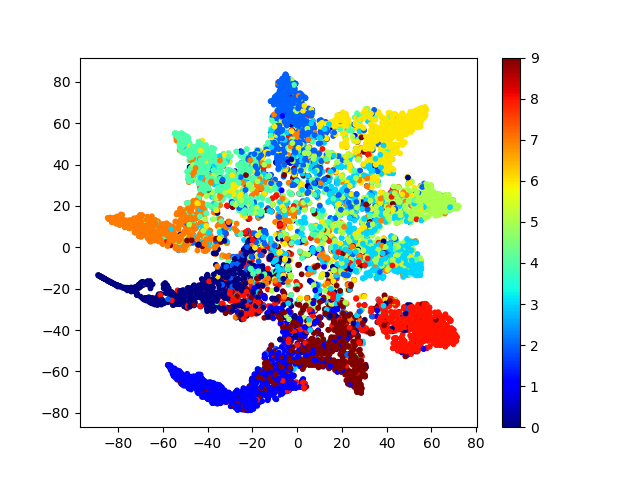}
    \end{minipage}%
    \begin{minipage}{0.24\hsize}
        \includegraphics[scale=0.3]{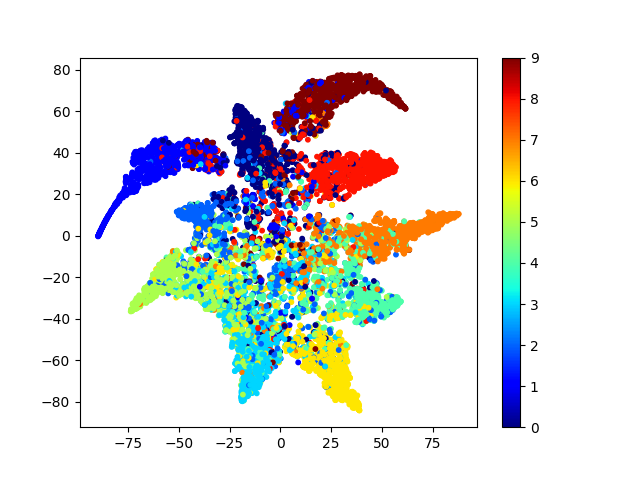}
    \end{minipage}%
    \begin{minipage}{0.24\hsize}
        \includegraphics[scale=0.3]{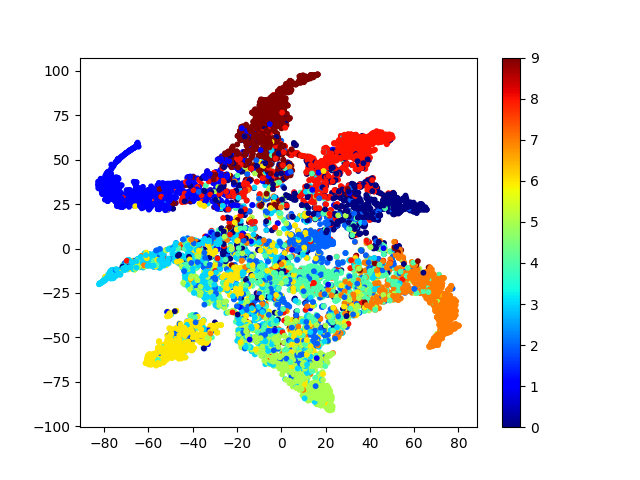}
    \end{minipage}%
    \end{tabular}
        \begin{tabular}{cccc}
      \begin{minipage}{0.24\hsize}
        \centering
        \includegraphics[scale=0.3]{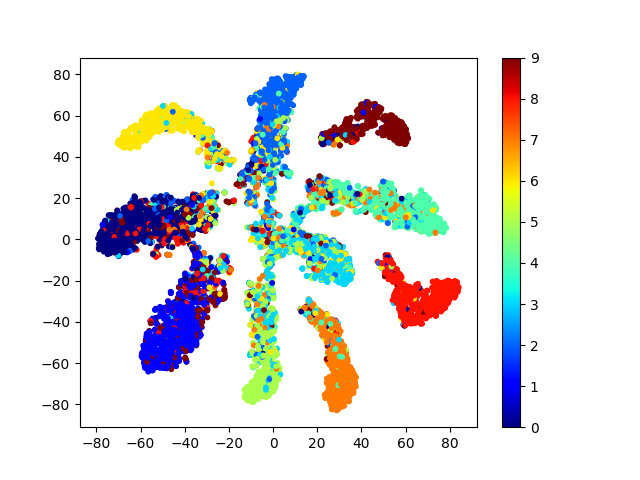}
        \subcaption{Long-tailed type(ration=5)}
    \end{minipage}%
    \begin{minipage}{0.24\hsize}
        \centering
        \includegraphics[scale=0.3]{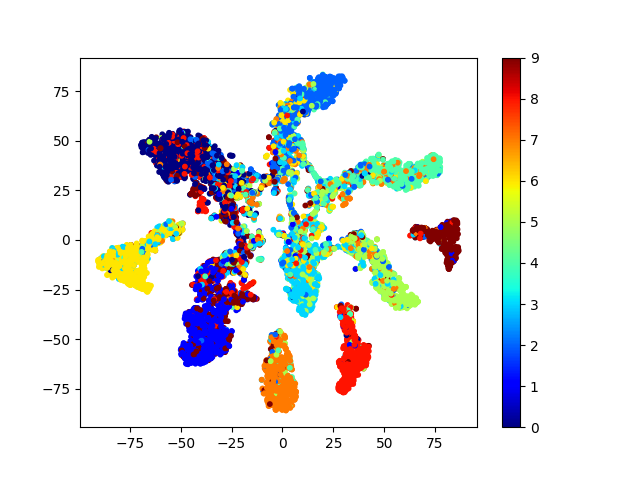}
        \subcaption{Long-tailed type(ration=10)}
    \end{minipage}%
    \begin{minipage}{0.24\hsize}
        \centering
        \includegraphics[scale=0.3]{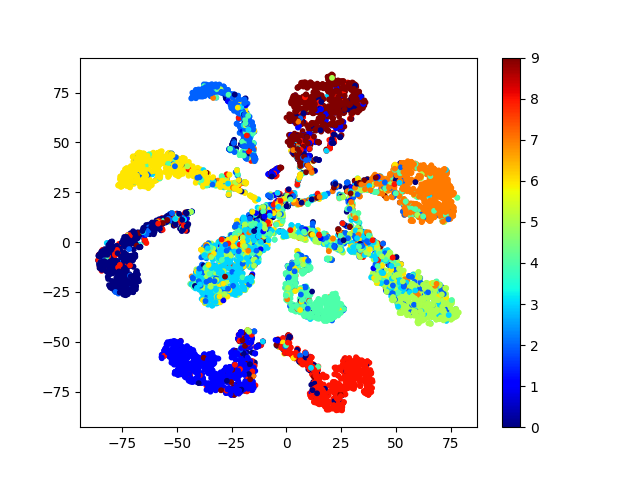}
        \subcaption{Step type(ration=5)}
    \end{minipage}%
    \begin{minipage}{0.24\hsize}
        \centering
        \includegraphics[scale=0.3]{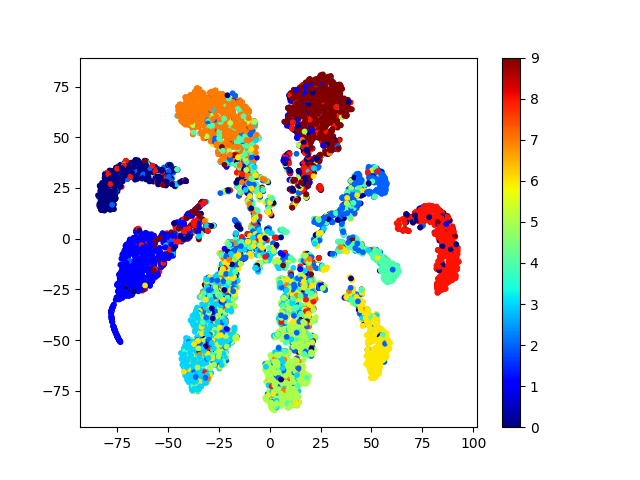}
        \subcaption{Step type(ration=10)}
    \end{minipage}%
    \end{tabular}
    \caption{Visualization of penultimate layer in ResNet34 trained on the imbalanced CIFAR-10. The first row shows the visualization results of CE loss and the second row shows the results of our method.}
\end{figure*}
\begin{figure*}[t]
    \centering
    \begin{tabular}{cccc}
      \begin{minipage}{0.24\hsize}
        \centering
        \includegraphics[scale=0.3]{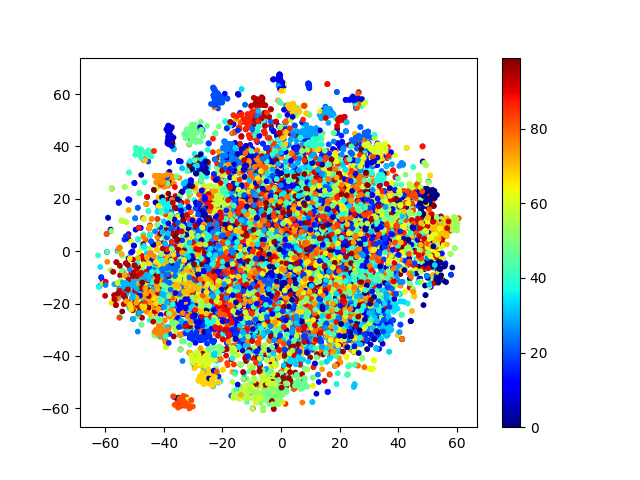}
    \end{minipage}%
    \begin{minipage}{0.24\hsize}
        \includegraphics[scale=0.3]{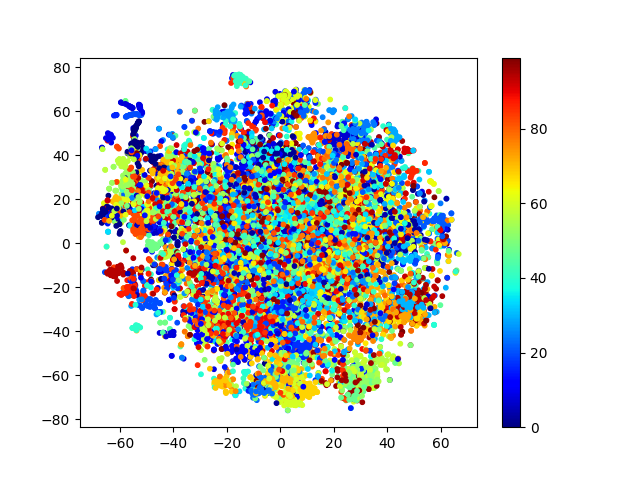}
    \end{minipage}%
    \begin{minipage}{0.24\hsize}
        \includegraphics[scale=0.3]{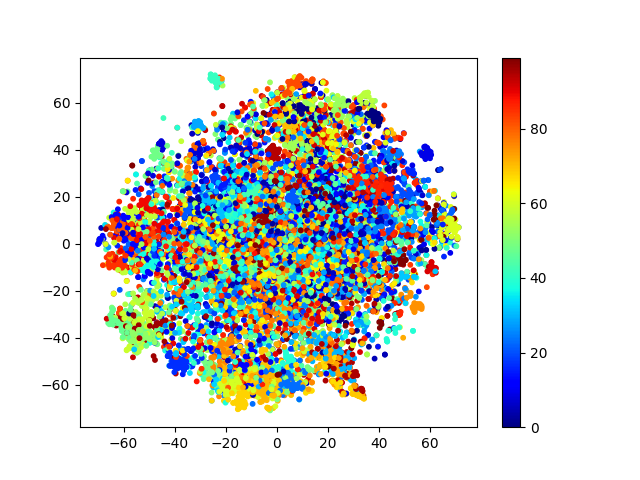}
    \end{minipage}%
    \begin{minipage}{0.24\hsize}
        \includegraphics[scale=0.3]{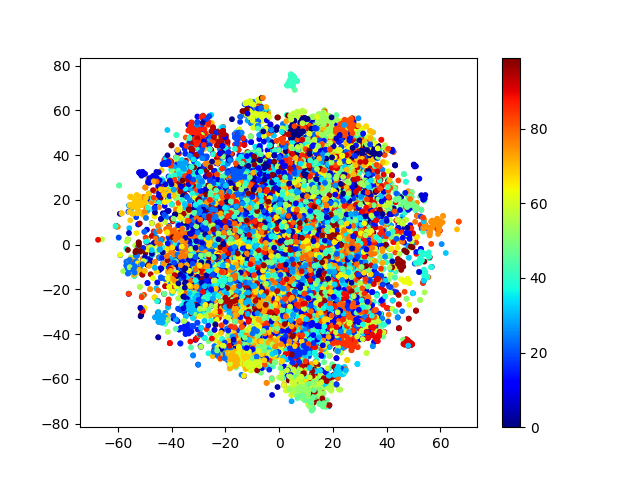}
    \end{minipage}%
    \end{tabular}
        \begin{tabular}{cccc}
      \begin{minipage}{0.24\hsize}
        \centering
        \includegraphics[scale=0.3]{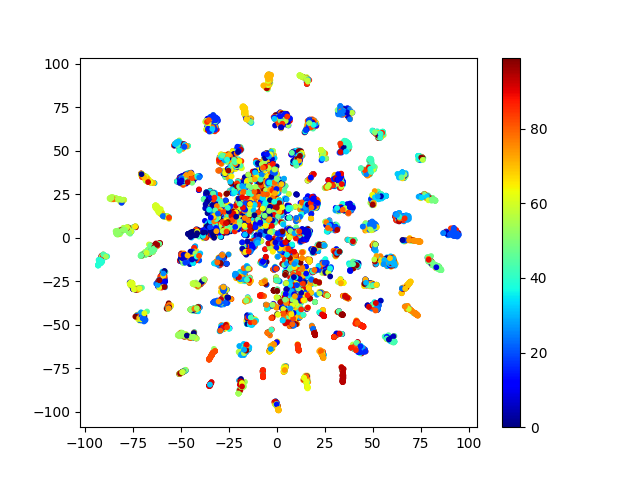}
        \subcaption{Long-tailed type(ration=5)}
    \end{minipage}%
    \begin{minipage}{0.24\hsize}
        \centering
        \includegraphics[scale=0.3]{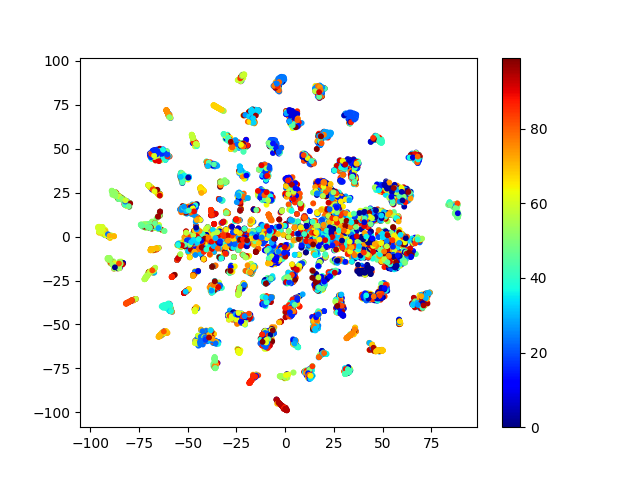}
        \subcaption{Long-tailed type(ration=10)}
    \end{minipage}%
    \begin{minipage}{0.24\hsize}
        \centering
        \includegraphics[scale=0.3]{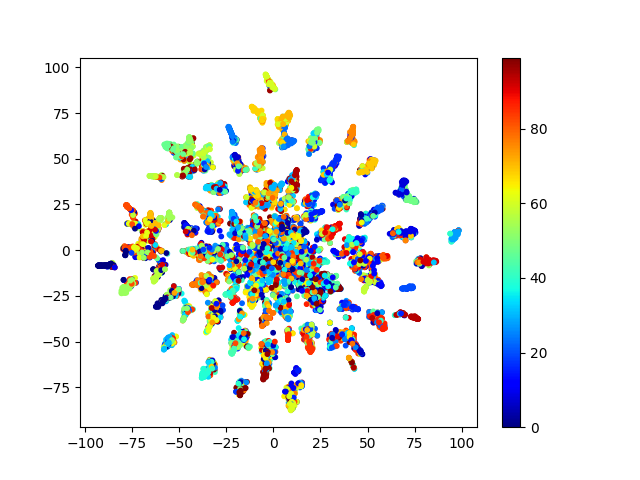}
        \subcaption{Step type(ration=5)}
    \end{minipage}%
    \begin{minipage}{0.24\hsize}
        \centering
        \includegraphics[scale=0.3]{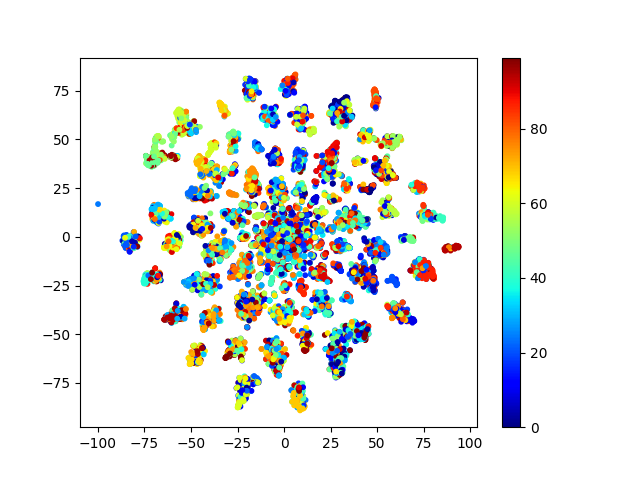}
        \subcaption{Step type(ration=10)}
    \end{minipage}%
    \end{tabular}
    \caption{Visualization of penultimate layer in ResNet34 trained on the imbalanced CIFAR-100. The first row shows visualization results of CE loss and the second row shows results of our method.}
\end{figure*}

\section{Overview of experiments}

To confirm the effectiveness of our proposed method, we conducted two kinds of experiments; imbalanced classification and semantic segmentation.
Section 4.1 shows the experiments on imbalanced classification and Section 4.2 shows the results on semantic segmentation.

\subsection{Imbalanced classification}

We conducted two kinds of experiments as the imbalanced classification.
The first experiments uses the CIFAR-10 and CIFAR-100 with pseudo imbalanced data. 
The second experiment uses long-tailed Food-101\cite{bossard2014food} data regarded as a middle-scale imbalanced dataset.

The original version of CIFAR-10 and CIFAR-100 contains 5,000 and 500 training images of  32×32 pixels with 10 and 100 classes.
To create imbalanced datasets, we reduced the number of training images and made two kinds of imbalanced datasets; long-tailed imbalance\cite{cui2019class} and step imbalance\cite{cao2019learning}.
We used imbalance ratio to denote between the samples of the most frequent and least frequent class and make four types of datasets with different imbalance ratios. 
Five images of each class selected randomly from training images were used as validation images.
Evaluation images were comprised of original test images.
We used ResNet34\cite{he2016deep} with full-scratch learning. 
The batch size for training was set to 128 and the optimization method was Adam({\sl $lr$}=$1\times10^{-3}$). We trained it till 1,000 epochs.
Experiments were conducted three times by changing the random initial values of the parameters, and the averaged accuracy of the three times experiments were used for evaluation. 

Food-101\cite{bossard2014food} contains the images of 101 food categories. 
Each category comprises 750 training and 250 evaluation images and provide 101,000 food images in total.
Similarly with the imbalanced CIFAR-10 and CIFAR-100, training images of Food-101 was made imbalanced using pseudo processing. We call it long-tailed Food-101 dataset.
This dataset composes 15,359 training images, 5,050 validation images and 25,250 test images in total and all images are 224×224 pixels.
Left column in Figure 5 shows the distribution of training samples in the long-tailed Food-101 dataset.
In experiment on the long-tailed Food-101 dataset, we used the ResNet50 pre-trained by the ImageNet\cite{krizhevsky2012imagenet} to investigate the effectiveness of our proposed MSE loss with the outlying label to the pre-trained model.
The batch size for training was set to 64 and SDG ($Momentum=0.9$) with weight decay was used as the optimization.
The learning rate $lr$ was decreased from the initial value $lr_{base}=1\times10^{-1}$ by using equation (6). We trained the network till 300 epochs.
\begin{equation}
  lr = lr_{base}\times\left(1-\frac{epoch}{epoch_{max}}\right)^{0.9} \\
\end{equation}

Classification accuracy and F-measure were used as the evaluation measures.
Accuracy is the measure how well the answer matches the overall prediction.
The F-measure is the harmonic mean of precision and recall.

\subsection{Semantic Segmentation}
We use the CamVid road scenes dataset (CamVid dataset)\cite{brostow2009semantic}, which is a dataset captured by vehicle video system, as the example of class imbalanced semantic segmentation. 
This dataset consists of 367 training images, 101 validation images and 233 evaluation images of 360×480 pixels. 
The challenge is to segment 11 classes such as road, building, cars, pedestrians, signs, poles, side-walk and so on.
The training images were augmented by random cropping of 320×320 pixels, horizon flipping, transforming brightness, contrast, and saturation. 
The right column in Figure 5 shows the distribution of training samples in the CamVid dataset.

FastFCN\cite{DBLP:journals/corr/abs-1903-11816} was used as the baseline network. Batch size during training was set to 16 and optimization method was SGD ($Momentum=0.9$) with weight decay. It was decreased learning rate $lr$ from the initial value $lr_{base}=1\times10^{-2}$ by using equation (6). 
We trained FastFCN till 200 epochs. 
The initial random number was changed three times and the averaged value of three times evaluations was used for evaluation. 
The evaluation measure was mean IoU (mIoU).

\section{Experiment results}
\subsection{Imbalanced CIFAR-10 and CIFAR-100}

Table 1 shows the accuracy on the imbalanced CIFAR-10, and Table 2 shows the F-measure.  Comparison results with conventional methods are also shown. 
In the Imbalanced CIFAR-10, except for the imbalance ratio=50, the accuracy of the proposed method (MSE Loss with OL) was the best in comparison with conventional methods based on CE loss.

When the imbalance ratio was 100, the accuracy of our method has improved 2.49\% and F-measure has improved 3.64\% for the long-tailed CIFAR-10 in comparison with CE loss.
Especially, since F-measure is the index involved with accuracy of each class, the improvement of F-measure is demonstrated the effectiveness of proposed method on the imbalanced datasets.

Table 3 and 4 show the accuracy and F-measure on the imbalanced CIFAR-100. 
Our proposed method (MSE Loss with OL) was the best in comparison with conventional methods in all imbalanced patterns. 
When the imbalance ratio was 100, the accuracy of our method has improved 4.87\% for the long-tailed CIFAR-100 and 3.27\% for the step type of CIFAR-100 in comparison with CE loss.
F-measure of our method has improved 3.64\% for the long-tailed type of CIFAR-100 and 0.79\% for the step type of CIFAR-100. 

\subsection{Visualization of penultimate layer}

Figure 6 and 7 show the visualization results of features at penultimate layer that compressed to two dimensions by t-SNE[6]. 
The first row are visualization results learned by CE loss and the second row are the results learned by our method.
In figures, left two columns show the results of long-tailed type ((a),(b)) and right two columns are the results of step type ((c),(d)).
Color bar shows the class number and indicates a small number of samples as the class number increase.
In the case of the imbalanced CIFAR-10, the distance between each class was small by CE loss. There were points where the features of other classes overlap near the center as shown in the first row of Figure 6.
It was confirmed that this phenomenon in all imbalanced patterns.
However, as shown in the second row of Figure 6, each class in the proposed method was the independent and it was possible to create the feature space for separating all classes. 
Since this feature space was separated high-difficulty classes and low-difficulty classes, the network prediction based on the separated features was prevented false prediction.

In the case of imbalanced CIFAR-100, each feature map was clearly difference.
Results of learning by CE loss, the features of each class were overlapped near the center in all imbalanced ratio as shown in the first row of Figure 7.
However, we confirmed that the relationship between each class was more clearly learned by the MSE loss with outlying labels.

\subsection{Long-tailed Food-101}
\begin{table}[t]
    \centering
    \caption{Top-1 evaluation accuracy on Long-tailed Food-101.}
    \scalebox{1.0}{
    \begin{tabular}{c||cc} \hline
    Method & Accuracy & F-measure\\
    \hline
        Weighted CE Loss\cite{cui2019class} & 50.98 & 48.11\\
        Focal Loss\cite{lin2017focal} & 41.52 & 38.02\\
        MSE Loss with OL($\alpha=1$) & \textbf{52.41} & \textbf{49.50}\\ 
    \hline
    \end{tabular}
    }
\end{table}

Table 5 shows the top-1 accuracy on the long-tailed Food-101 dataset. 
Our proposed method has improved 1.43\% in comparison with the weighted CE loss.
Although long-tailed Food-101 has almost the same number of classes as CIFAR-100, it was more difficult than CIFAR-100 because it was close to the real-world dataset.
Accuracy improvement on this dataset shows that the proposed method was also effective for the real-world dataset.

\subsection{CamVid}
\begin{table}[t]
    \centering
    \caption{Mean IoU on test images in the CamVid. We compared our method with conventional loss functions for semantic segmentation.}
    \scalebox{1.1}{
    \begin{tabular}{c||c} \hline
    Method & mIoU \\ 
    \hline
        Weighted CE Loss\cite{badrinarayanan2017segnet} & 67.49(±0.14) \\
        Weighted Focal Loss\cite{lin2017focal} & 52.94(±0.06) \\
        Dice Loss\cite{milletari2016v} & 69.75(±0.36) \\ 
        IoU Loss\cite{rahman2016optimizing} & 67.05(±0.06) \\
        Tversky Loss\cite{salehi2017tversky} & 44.10(±0.40) \\
        Lovasz-hinge Loss\cite{berman2018lovasz} & 68.54(±0.50) \\ 
        MSE Loss with OL($\alpha=6$) & \textbf{72.69(±0.19)} \\ 
    \hline
    \end{tabular}
    }
\end{table}
\begin{table}[t]
    \centering
    \caption{Comparison of mIoU while changing $\alpha$. The $\alpha$ was determined by the validation dataset.}
    \scalebox{1.2}{
    \begin{tabular}{c||c} \hline
    Method & mIoU \\ 
    \hline
        $\alpha=1$ & 69.49(±0.04) \\
        $\alpha=2$ & 71.56(±0.14) \\
        $\alpha=3$ & 71.90(±0.20) \\ 
        $\alpha=4$ & 72.41(±0.22) \\
        $\alpha=5$ & 72.01(±0.42) \\
        $\alpha=6$ & \textbf{72.69(±0.19)} \\ 
        $\alpha=7$ & 71.80(±0.37) \\
        $\alpha=8$ & 70.63(±0.79) \\
    \hline
    \end{tabular}
    }
\end{table}


Table 6 shows the mIoU on test images in the CamVid dataset.
We evaluated the weighted CE loss\cite{badrinarayanan2017segnet}, weighted Focal loss\cite{lin2017focal}, Dice loss\cite{milletari2016v}, IoU loss\cite{rahman2016optimizing}, Tversky loss\cite{salehi2017tversky} and Lovasz-hinge loss\cite{berman2018lovasz} as the conventional methods for semantic segmentation.
When our proposed the MSE loss with OL was used, mIoU has achieved 72.69\% and the improvement was 5.2\% in comparison with the weighted CE loss. 
Note that the optimal $\alpha$ is determined by validation dataset.
Although Dice loss was the best mIoU among the conventional loss functions, the proposed method has improved by 2.94\%. 
We demonstrated the superiority of the proposed method to conventional loss functions for semantic segmentation.

Table 7 shows the mIoU while changing the hyper-parameter $\alpha$ from 1 to 8.
By increasing the value of $\alpha$, the difference between the classes with a large number of samples and the classes with a small number of samples became large.
The best mIoU was obtained at $\alpha=6$ and  mIoU has ahiceved 72.69\%.
From Table 6 and 7, the mIoU of our method has improved 2.00\% in comparison with weighted CE loss even if we use $\alpha=1$.

Figure 8 shows the visualization result of penultimate layer in the FastFCN. 
When we increased $\alpha$, we confirmed that wine-red class ("Bicyclist") is placed more outside from the center. The classes with skyblue ("Side-walk"), red ("Pedesrian") and dark orange ("Car") with the small number of samples were more independent.
Figure shows that adequate value of $\alpha$ creates better feature space.
Throughout all experiments, we demonstrated the effectiveness of our method for imbalanced data.

\begin{figure}
    \centering
    \begin{tabular}{cc}
      \begin{minipage}{0.45\hsize}
        \centering
        \includegraphics[scale=0.26]{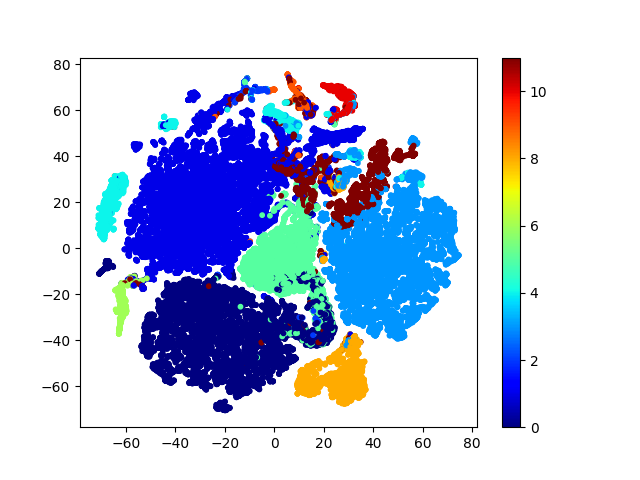}
    \end{minipage}%
    \begin{minipage}{0.45\hsize}
        \centering
        \includegraphics[scale=0.26]{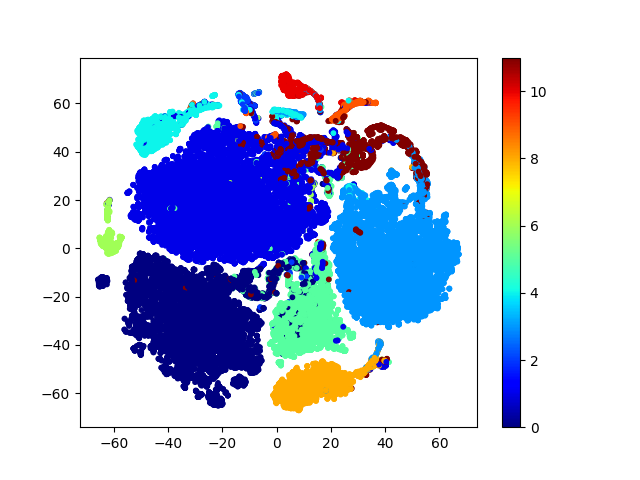}
    \end{minipage}%
    \end{tabular}
    \caption{Visualization of penultimate layer’s activations of FastFCN on the CamVid dataset by t-SNE\cite{van2008visualizing}. 
    The left column shows the visualization of $\alpha=1$, and the right column shows the visualization of $\alpha=6$.
    }
\end{figure}

\section{Conclusion}


In this paper, we proposed the MSE loss with outlying label for imbalanced datasets. 
By the experiments on imbalanced classification and semantic segmentation, the MSE loss with outlying label significantly improved the accuracy and F-measure in comparison with conventional methods based on CE loss.
Furthermore, we confirmed that the proposed method created the effective feature space for classifcation.

However, $\alpha$ is the hyper-parameter that we must determine, and we need to conduct some experiments to set the optimal $\alpha$. 
How to set $\alpha$ without experiments using distribution of training images is one of the future work.


{\small
\bibliographystyle{ieee_fullname}
\bibliography{egbib}
}

\end{document}